# Image Labeling and Segmentation using Hierarchical Conditional Random Field Model

Mr. Manoj K.Vairalkar[1] and Mrs.Sonali.Nimbhorkar[2]

[1]Department of Computer Science and Engineering, RTMN University,Nagpur

mkvairalkar@gmail.com

[2] Department of Computer Science and Engineering, RTMN University,Nagpur

nimsonali21@yahoo.com

## ABSTRACT

*The use of hierarchical Conditional Random Field model deal with the problem of labeling images . At the time of labeling a new image, selection of the nearest cluster and using the related CRF model to label this image. When one give input image, one first use the CRF model to get initial pixel labels then finding the cluster with most similar images. Then at last relabeling the input image by the CRF model associated with this cluster. This paper presents a approach to label and segment specific image having correct information.*

## 1. INTRODUCTION

Segmentation is the decomposition of an image into these objects and regions by associating every pixel with the object that it corresponds to. Most humans can easily segment an image. Computer automated segmentation is a difficult problem, requiring sophisticated algorithms that work in tandem.

labeling identifies an object record of an information page based on the labeling of object elements within an object record and labels object elements based on the identification of an object record that contains the object elements.

The understanding of image and object identification is the core technology. Here the goal is to assign every pixel of the image with an object class label. For this the solutions fall into two general categories: parametric methods and nonparametric methods.

Parametric methods [1] mostly rivet minimizing a Conditional Random Field (CRF) model to assign a particular label to every pixel on the basis of probability. In learning phase of parametric method, the parameters of the CRF models are optimized with the help of training examples, and in an inference phase, the CRF model is useful to label a test image. In difference to parametric methods, nonparametric methods [10] do not involve any training. The design of these methods is to relocate labels from a recovery set which contains meaningful similar images. Here, the approach is to label specific type of image whose correct information is present.

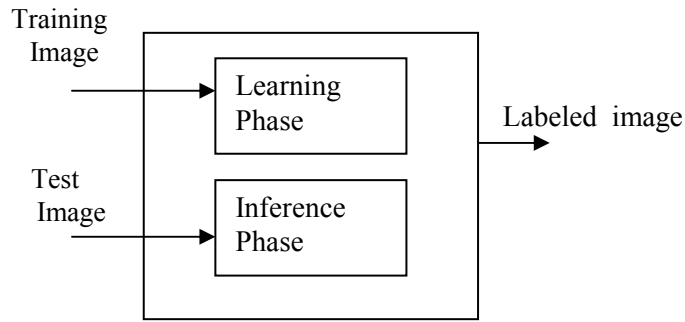
Fig.1: Conditional Random Field Model

In learning phase, firstly the training of image to the system carried out then according to the test image type, the inference phase tries to matches input image with trained system on the basis of probability.

## 2. RELATED WORK

### 2.1. Parametric methods

Image labeling by optimizing a CRF model has proven to be the state-of-the-art parametric image labeling method. Traditional CRF models combine unary energy terms, which evaluate the possibility of a single pixel taking a particular label, and pair-wise energy terms, which evaluate the probability of adjacent pixels taking different labels. Although these approaches work well in many cases, they still have their own limitations because these CRF models are only up to second-order and it is difficult to incorporate large-scale contextual information.

### 2.1. Nonparametric methods

The key components of nonparametric methods are how to find the retrieval set which contains similar images, and how to build pixel-wise or super pixel-wise links between the input image and images in the retrieval set.

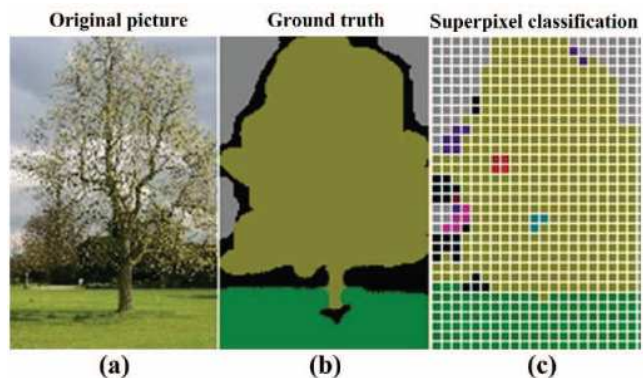
Fig.2: Illustration of super pixel classification

# 3. MODELING METHODS

## 3.1. Label Based Descriptor

By labeled image, each pixel is labeled as one of the $k$ object classes $L = \{m_k\}$. The meaningful data of an image is obtained from the position, appearance and shape of each object in the image. Here, the value of can be varied.

## 3.2. Positional information

The positional information can be obtained by dividing image, grid. When we divide image it forms cell formulation. Within each grid cell say $g_{ij}$, calculate the distribution $S_{ijk}$ of every object class $m_k \in L$. Then assemble all the cell coverage information into a vector $d^pI$ of length $K(np)^2$. A big $np$ would make $d^pI$ capture the positional information more precisely, while a small $np$ would make $dp\ I$ less sensitive to image displacement and classification errors. If L contains 3 objects and grid value is 3 then the calculation yields to 27. If L contains 3 objects then the representation is
$L=\{m_0,m_1,m_2\}$
Where $m_0$ can be treated as object1, $m_1$ can be represented as object2 and so on
The partitioning of 9 x 9 image represented as:

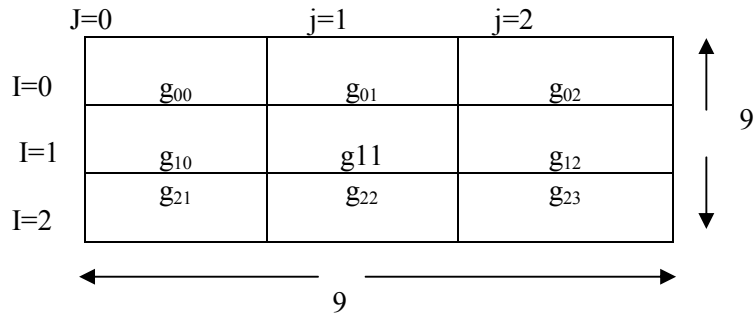

## 3.3 Appearance

The appearance information obtained by calculating the mean color $c_{ijk} = (r_{ijk}, g_{ijk}, b_{ijk})$ of each object class $c_k$ within each cell $g_{ij}$. To stabilize the mean color statistics, scale each mean color $c_{ijk}$ as $p_{ijk} c_{ijk}$. Again, all mean colors $c_{ijk}$ are collected into a vector $d^cI$ of length $3K(np)2$.

# 4. IMAGE LOADING

In image loading module, the Java Advanced Imaging Application Program Interface is used that allows reading and writing of various image formats, and helps for manipulation. Then getting the Red, Green and Blue value for given image one is able to binaries the image.

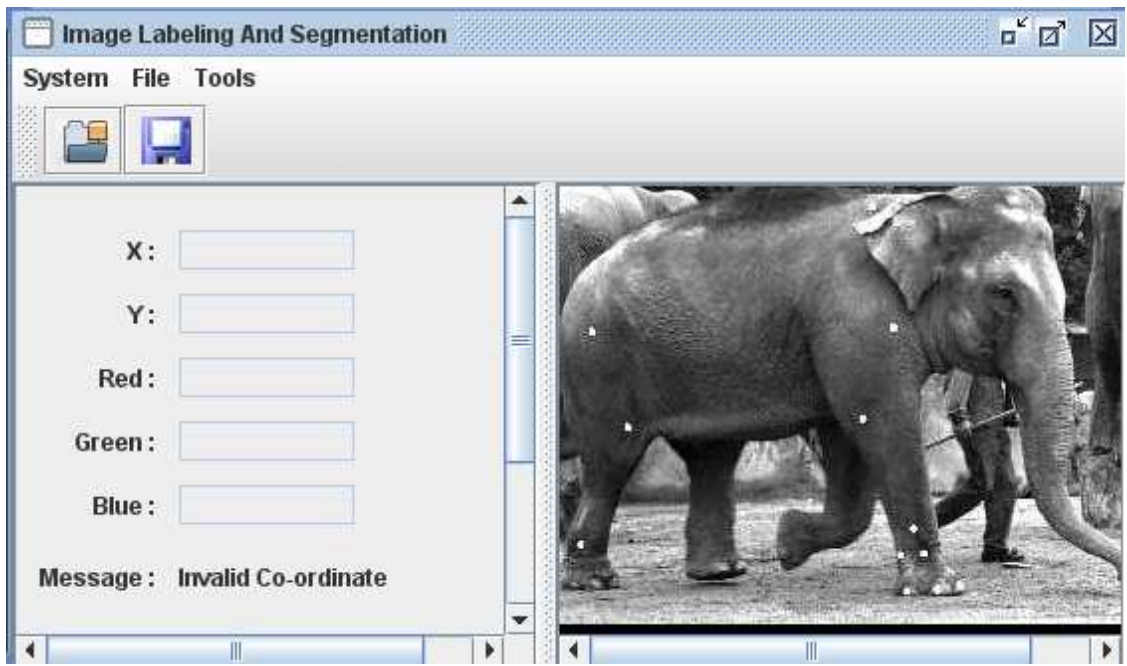

Fig.3: Image Loading Sample

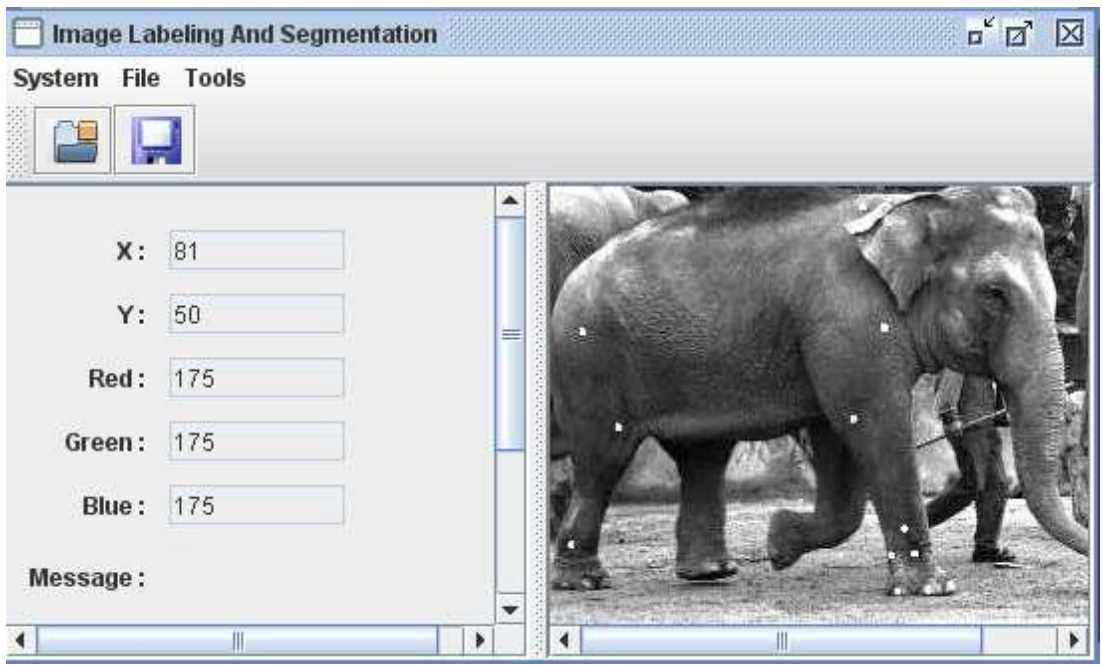

Fig.4: Image Binarization

The flow of process of image loading and binarization shown as follows.

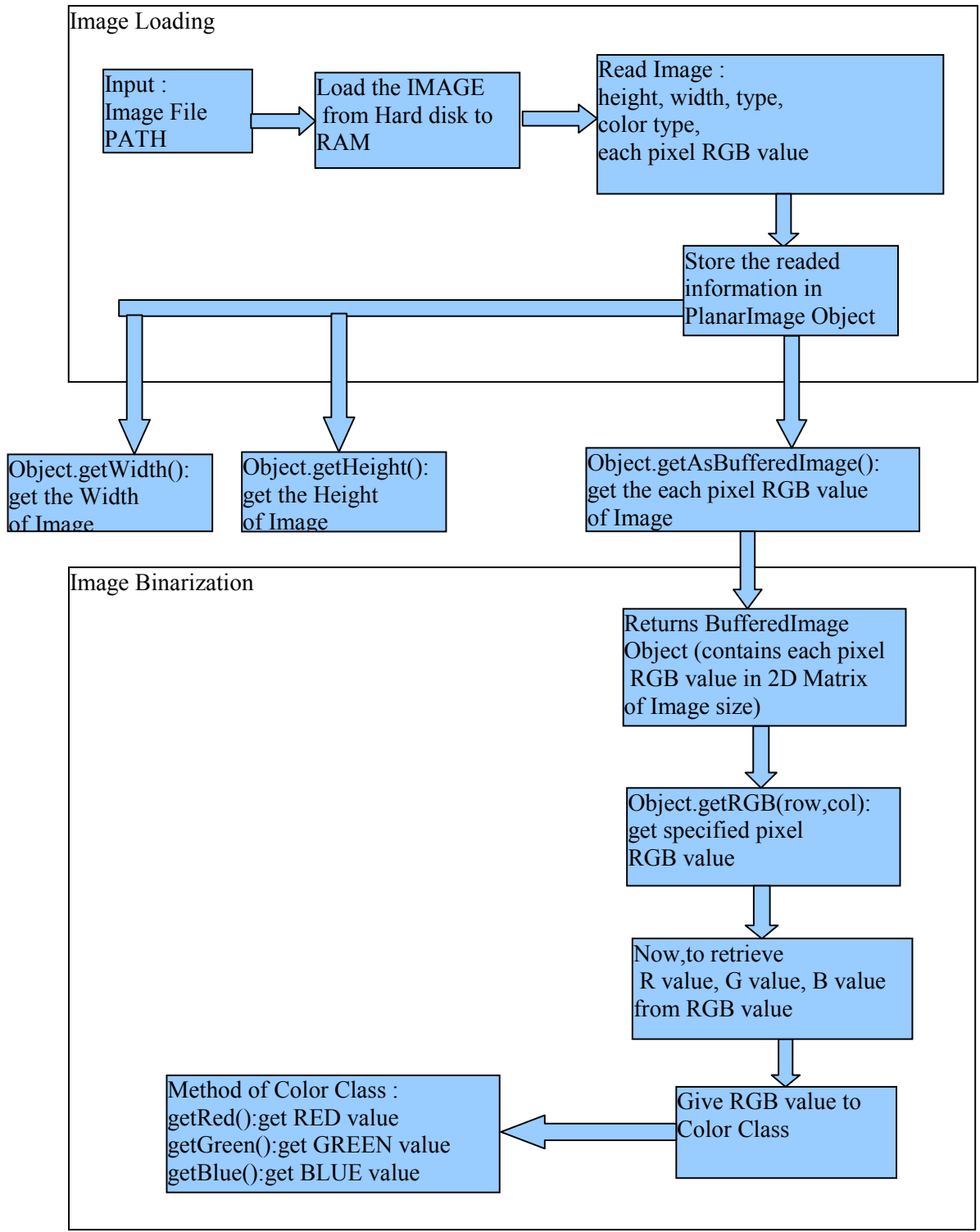

Fig.5: Block Diagram for Image Loading and Binarization

## 5. SEGMENTATION

Segmentation refers to the practice of partition a digital image into various segments. The objective of segmentation is to simplify and/or modify the depiction of an image into something that is more significant and easier to analyze. Image segmentation is used to locate objects and boundaries (lines, curves, etc.) in images. More precisely, image segmentation is the process of assigning a label to every pixel in an image such that pixels with the same label share certain image characteristics.

The result of image segmentation is a set of segments that collectively cover the entire image, or a set of contours extracted from the image. Each of the pixels in a area are like with respect to some characteristic such as color, intensity, or texture. Color image segmentation is useful in many applications. From the segmentation results, it is possible to identify regions of interest and objects in the scene, which is very beneficial to the successive image analysis.

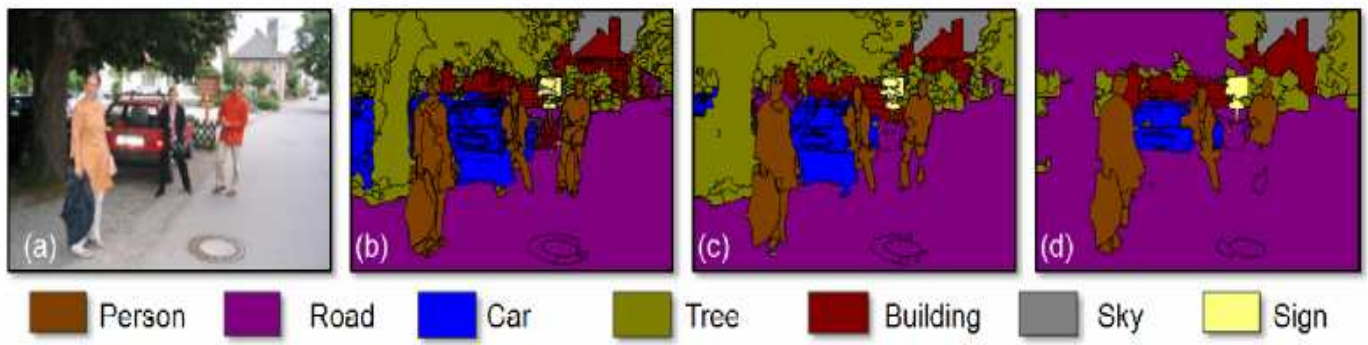

Fig.6.Image Labeling and segmentation

## 6. WAVELET TRANSFORM

The optimization of speed performance can be achieved with the help of wavelet transform as compression is possible. Wavelets are mathematical functions that scratch up data into unlike frequency components, and then study each component with a resolution matched to its scale. There are advantages over traditional Fourier methods in analysing physical situations where the signal contains discontinuities and sharp spikes. Wavelets were developed independently in the fields of mathematics, quantum physics, and electrical engineering. Wavelets are functions that satisfy certain mathematical requirements and are used in representing data or other functions. This idea is not new. Approximation using superposition of functions The wavelet transform is a mathematical tool that decomposes a signal into a representation that shows signal details and trend as a function of time. One use this representation to characterize temporary events, decrease noise, compress data, and perform many other operations. The main advantage of wavelet methods over traditional Fourier methods are the use of localized basis functions and the faster computation speed. Localized basis functions are ideal for analysing real material situations in which a signal contains discontinuities and sharp spikes. Wavelet transform is an evolving technology which offers far higher degrees of data compression compared to standard transforms

## 7. CONCLUSION

In this paper, presents an approach to label the images whose correct information is present. The Hierarchical CRF model finds similar object from trained examples and assign suitable priority according to object to label. The experimental results looks as figure (1) naming Effect of image quantisation on object segmentation where (a) Original image. (b)-(d) Object class segmentations with different images. (b), (c) and (d) use three different unverified segmentations of the image. Each segment is assigned the label of the dominant object present in it. It can be seen that (b) is the best for tree, road, and car. However, (d) is better for the left person and the sign board.

**Authors**

**M.Vairalkar** received the B.E.. degree in Information Technology from RTMN University, Nagpur, India in 2006 and pursuing M.Tech degree in Computer Science and Engineering from RTMN University,Nagpur, India. Attended the 2[nd] International Conference On Emerging Trends in Engineering and Technology-09 at Nagpur. He has a Life Member ship in ISTE.

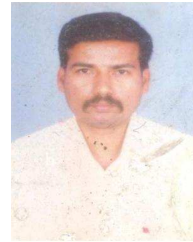

**S.Nimbhorkar** received the M.Tech degree in Computer Science and Engineering from RTMN University , India She is currently working as an Assistant professor in Computer Science and Engineering Department in GHRCE Nagpur, Maharashtra, India. She has a Life Member ship in ISTE. She published 3 International Research papers.

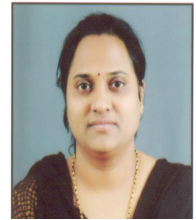